\pdfoutput=1

\documentclass[11pt]{article}

\usepackage{acl}

\usepackage{times}
\usepackage{soul}
\usepackage{url}
\usepackage[utf8]{inputenc}
\usepackage{graphicx}
\usepackage{amsmath}
\usepackage{amsthm}
\usepackage{booktabs}
\usepackage{algorithm}
\usepackage{algorithmic}
\usepackage{subfigure}
\usepackage{amssymb}
\usepackage{amsmath}
\usepackage{multirow}
\usepackage{booktabs}
\usepackage{array}
\usepackage{arydshln}
\usepackage{pgfplots, pgfplotstable}
\usepackage{caption}
\usepgfplotslibrary{groupplots}
\usepackage{tikz}
\usepackage{bm}
\usepackage{CJKutf8}
\usepackage{color}

\usepackage[T1]{fontenc}

\usepackage{microtype}

\def\newcite#1{\citeauthor{#1}~\shortcite{#1}}

\makeatletter
\def\adl@drawiv#1#2#3{%
        \hskip.5\tabcolsep
        \xleaders#3{#2.5\@tempdimb #1{1}#2.5\@tempdimb}%
                #2\z@ plus1fil minus1fil\relax
        \hskip.5\tabcolsep}
\newcommand{\cdashlinelr}[1]{%
  \noalign{\vskip\aboverulesep
           \global\let\@dashdrawstore\adl@draw
           \global\let\adl@draw\adl@drawiv}
  \cdashline{#1}
  \noalign{\global\let\adl@draw\@dashdrawstore
           \vskip\belowrulesep}}
\makeatother

\author{}

\title{OneEE: A One-Stage Framework for Fast Overlapping and Nested \\Event Extraction}

\author{
    Hu Cao\textsuperscript{\rm 1}$^*$,
    Jingye Li\textsuperscript{\rm 1}$^*$,
    Fangfang Su\textsuperscript{\rm 1},
    Fei Li\textsuperscript{\rm 1},
    Hao Fei\textsuperscript{\rm 2},
    Shengqiong Wu\textsuperscript{\rm 2},\\
    \textbf{
    Bobo Li\textsuperscript{\rm 1},
    Liang Zhao\textsuperscript{\rm 3},
    Donghong Ji\textsuperscript{\rm 1}$^\dagger$
    }\\
\textsuperscript{\rm 1} Key Laboratory of Aerospace Information Security and Trusted Computing, Ministry  \\of Education,
School of Cyber Science and Engineering, Wuhan University, China \\
\textsuperscript{\rm 2} School of Computing, National University of Singapore, Singapore\\
\textsuperscript{\rm 3} Department of Computing and Mathematics, University of São Paulo, Brazil\\
\texttt{\{whucaohu, theodorelee, lifei\_csnlp, dhji\}@whu.edu.cn} 
}

\begin{document}
\begin{CJK}{UTF8}{gbsn}

\maketitle
\renewcommand{\thefootnote}{\fnsymbol{footnote}}
\footnotetext[1]{Equal contribution}
\footnotetext[2]{Corresponding author}
\renewcommand{\thefootnote}{\arabic{footnote}}

\begin{abstract}
Event extraction (EE) is an essential task of information extraction, which aims to extract structured event information from unstructured text.
Most prior work focuses on extracting flat events while neglecting overlapped or nested ones.
A few models for overlapped and nested EE includes several successive stages to extract event triggers and arguments,
which suffer from error propagation.
Therefore, we design a simple yet effective tagging scheme and model to formulate EE as word-word relation recognition, called OneEE.
The relations between trigger or argument words
are simultaneously recognized in one stage with parallel grid tagging, thus yielding a very fast event extraction speed.
The model is equipped with an adaptive event fusion module to generate event-aware representations and a distance-aware predictor to integrate relative distance information for word-word relation recognition, which are empirically demonstrated to be effective mechanisms.
Experiments on 3 overlapped and nested EE benchmarks, namely FewFC, Genia11, and Genia13,
show that OneEE achieves the state-of-the-art (SoTA) results.
Moreover, the inference speed of OneEE is faster than those of baselines in the same condition, and can be further substantially improved since it supports parallel inference.\footnote{The codes at \url{https://github.com/Cao-Hu/OneEE}}
\end{abstract}

\section{Introduction}

\begin{figure}[!t]
    \centering
    \includegraphics[width=0.99\columnwidth]{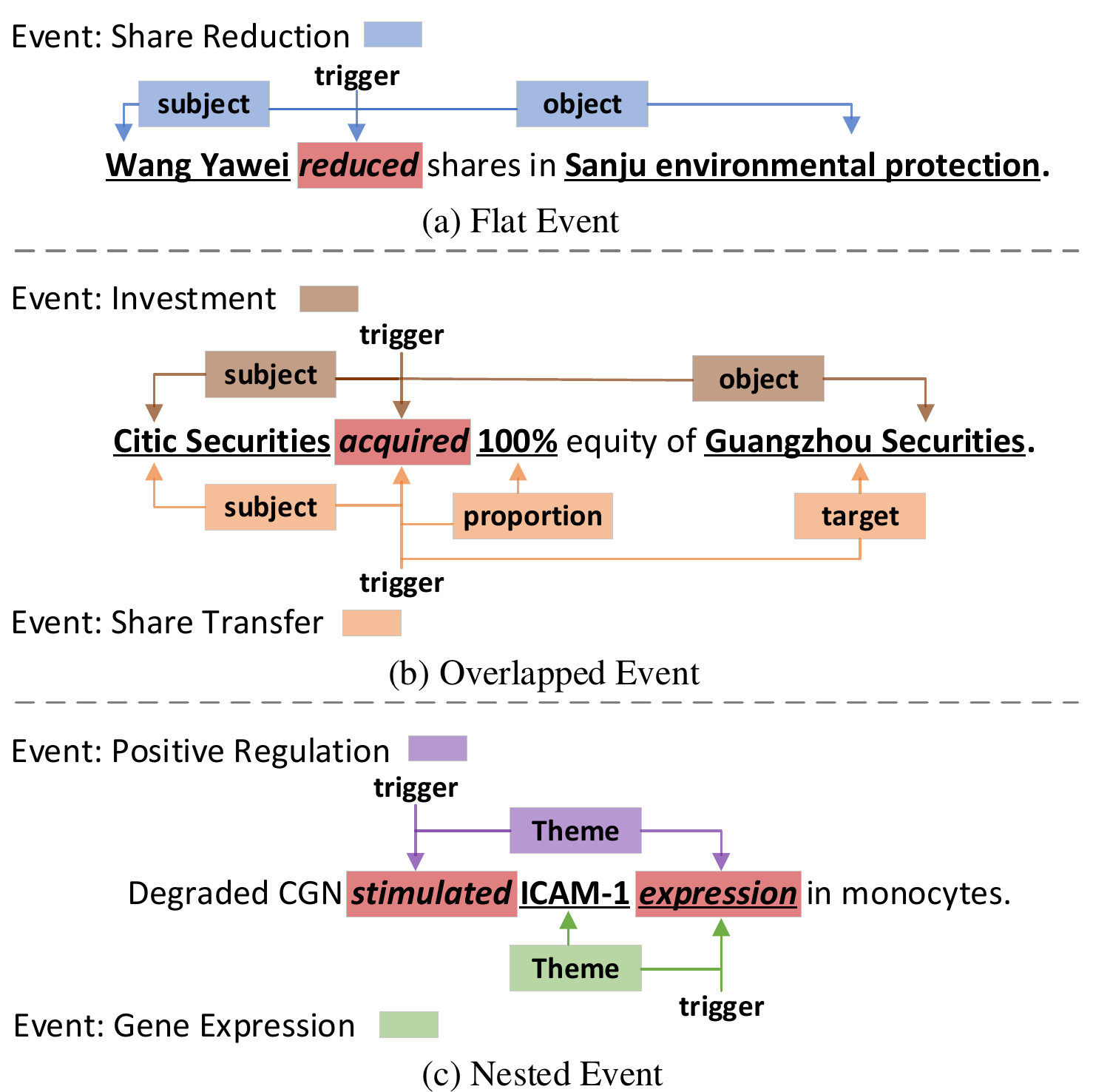}
    \caption{
    Examples of three kinds of events, including a flat event (a), overlapped events (b), and nested events (c).
    Different event mentions are denoted in distinct colors.
    Triggers are marked with red boxes while arguments are underlined.
    }
    \label{fig:example}
\end{figure}

Event Extraction (EE) is a fundamental yet challenging task in information extraction research \cite{miwa-bansal-2016-end,katiyar-cardie-2016-investigating,fei-etal-2020-cross,li-etal-2021-mrn,Feiijcai22DiaRE}. 
EE facilitates the development of practical applications such as knowledge graph construction \cite{MMGCN,bosselut2021dynamic}, biological process analysis \cite{miwa2013method}, and financial market surveillance \cite{nuij2013automated}.
The goal of EE is to recognize event triggers as well as the associated arguments from texts. 
As an example, Figure \ref{fig:example}(a) illustrates a \texttt{Share Reduction} event including a trigger ``reduced'' and a subject argument ``Wang Yawei''.

Traditional methods for EE \cite{li2013joint,chen2015event,nguyen2016joint,liu2018jointly,nguyen2019one} regard event extraction as a sequence labeling task, assuming that event mentions do not overlap with each other.
However, they neglect complicated irregular EE scenarios (i.e., overlapped and nested EE) \cite{fei2020boundaries,0001JLLRL21}.
As exemplified in Figure \ref{fig:example}(b), 
there are two overlapped events, \texttt{Investment}, and \texttt{Share Transfer}, which share the same trigger word ``acquired'' and the argument words ``Guangzhou Securities''.
Figure \ref{fig:example}(c) illustrates an example of nested events where the event \texttt{Gene Expression} is the \texttt{Theme} argument of another event \texttt{Positive Regulation}.

Prior studies for overlapped and nested EE \cite{yang2019exploring,li2020event} employ pipeline-based methods that extract event triggers and arguments in several successive stages.
Recently, the state-of-the-art model \newcite{sheng2021casee} also uses such a method that consecutively performs event type detection, trigger extraction, and argument extraction.
The main problem with such a method is that the latter stage relies on the former stage, which inherently brings the error propagation problem.

\begin{figure}[!t]
     \centering
     \subfigure[Event: Investment]{
         \centering
         \includegraphics[width=1\columnwidth]{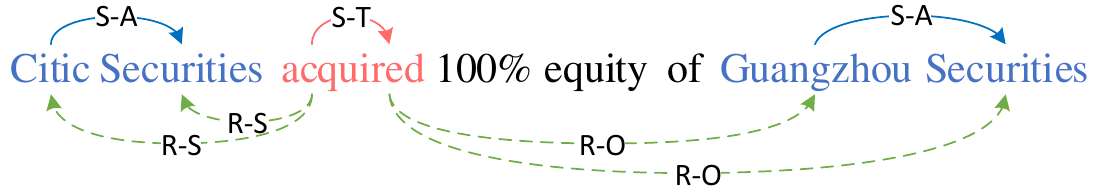}
         \label{fig:label1}
     }
     \subfigure[Event: Transfer Share]{
         \centering
         \includegraphics[width=1\columnwidth]{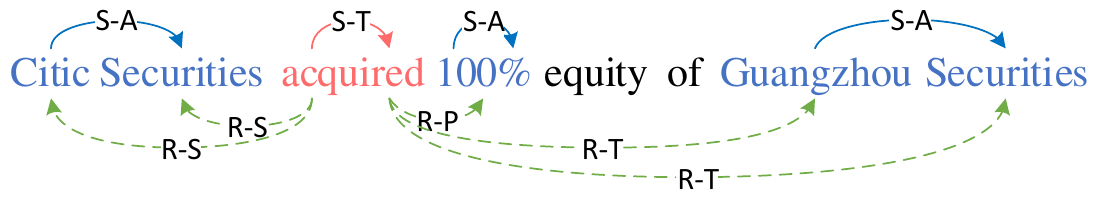}
         \label{fig:label2}
     }
    \caption{Two examples to illustrate our tagging scheme.
    We formalize the overlapping and nested EE as word-word relation recognition, where S-T and S-A denote the relations between the head and tail boundary words of a trigger or argument,
    and R-S, R-O, R-T, and R-P denote the relations between the trigger word and the argument words with the roles ``subject'', ``object'', ``target'' and ``proportion''.
    }
    \label{fig:label}
\end{figure}

To address the above issue,  we present a novel tagging scheme that transforms overlapping and nested EE into word-word relation recognition.
As shown in Figure \ref{fig:label}, we design two types of relations, including the span relation (S-\texttt{*}) and role relation (R-\texttt{*}).
S-\texttt{*} handles trigger and argument identification, denoting whether two words are the head-tail boundary of a trigger (T) or argument (A).
R-\texttt{*} addresses argument role classification, indicating whether the argument plays the ``\texttt{*}'' role in the event.

Based on this scheme, we further propose a one-stage event extraction model, OneEE, which mainly includes three parts.
First, it adopts BERT \cite{devlin2019bert} as the encoder to get contextualized word representations.
Afterward, an adaptive event fusion layer composed of an attention module and two gate fusion modules are used to obtain event-aware contextual representations for each event type.
In the prediction layer, we parallelly predict the span and role relations between each pair of words by calculating distance-aware scores.
Finally, event triggers, arguments, and their roles can be decoded out using these relation labels in one stage without error propagation.

We evaluate OneEE on 3 overlapped and nested EE datasets (FewFC \cite{zhou2021role}, Genia11 \cite{kim2011overview}, and Genia13 \cite{kim2013Genia}), and conduct extensive experiments and analyses.
Our contributions can be summarized as follows:

$\bullet$ We design a new tagging scheme that casts event extraction as a word-word relation recognition task, providing a novel and simple solution for overlapped and nested EE.

$\bullet$ We propose OneEE, a one-stage model that effectively extracts word-word relations in parallel for overlapped and nested EE.

$\bullet$ We further present an adaptive event fusion layer to obtain event-aware contextual representations and effectively integrate event information.

$\bullet$ OneEE outperforms the SoTA model with regard to both the performance and inference speed.

\begin{figure*}[ht]
    \centering
    \includegraphics[width=1\linewidth]{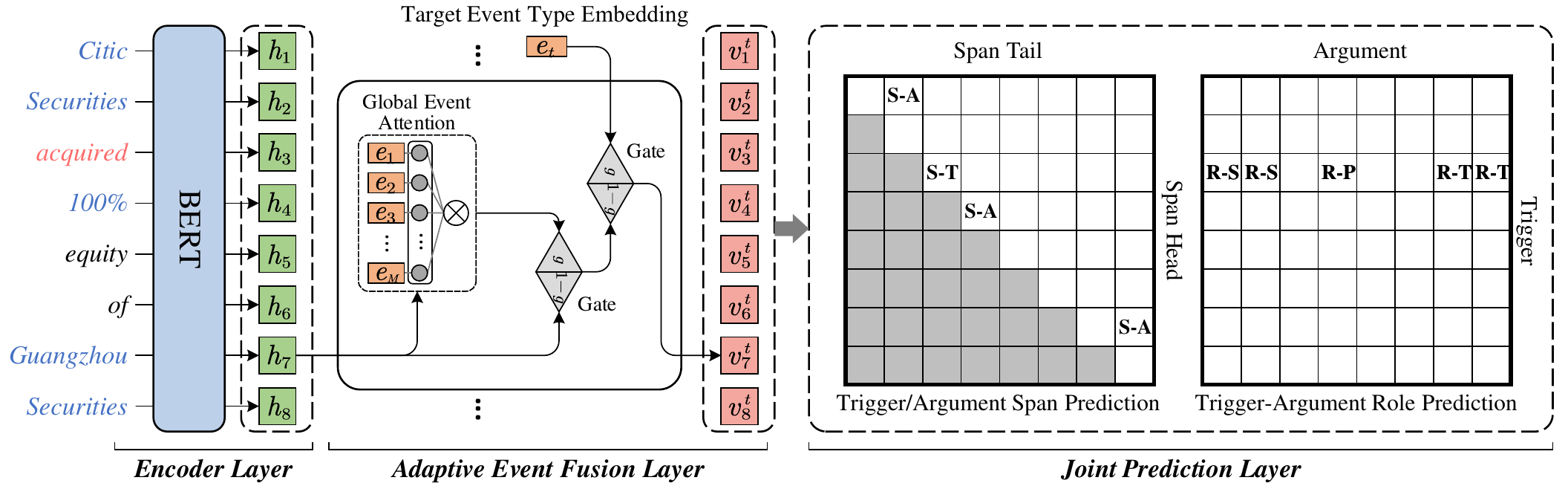}
    \caption{The architecture of our framework. Given a target event type embedding $\bm{e}_t$ of type $t$ (e.g., transfer share), the goal of our framework is to identify its triggers, arguments, and corresponding roles in the input sentence.}
    \label{fig:architecture}
\end{figure*}

\section{Related Work}

\subsection{Event Extraction}

Information extraction is one of the key research track in natural language processing \cite{miwa-bansal-2016-end,0001ZLJ21}, among which the event extraction is the most complicated task \cite{chen2015event,Feiijcai22DiaSRL}.
Traditional EE (i.e., flat or regular EE) \cite{li2013joint,nguyen2016joint,liu2018jointly,sha2018jointly,nguyen2019one} formulates EE into a sequence labeling task, assigning each token with a label (e.g., \textit{BIO} tagging scheme).
For example, \newcite{nguyen2016joint} uses two bidirectional RNNs to get richer representation which is then utilized to predict event triggers and argument roles jointly.
\newcite{liu2018jointly} jointly extracts multiple event triggers and arguments by introducing attention-based GCN to model the dependency graph information \cite{0001LLJ21,li-etal-2021-span,0001WRZ22}.
However, their underlying assumption that event mentions do not overlap with each other is not always valid.
Irregular EE (i.e., overlapped and nested EE) has not received much attention, which is more challenging and realistic.

Existing methods for overlapped and nested EE \cite{yang2019exploring,li2020event} perform event extraction in a pipeline manner with several steps.
To solve the argument overlap, \newcite{yang2019exploring} adopts multiple sets of binary classifiers where each severs for a role to detect the role-specific argument spans but fails in solving trigger overlap.
Except for pipeline methods, the latest attempt dealing with overlapped EE is \newcite{sheng2021casee} in a joint framework with cascade decoding.
They are the first to simultaneously tackle all the overlapping patterns.
\newcite{sheng2021casee} sequentially performs type detection, trigger extraction, and argument extraction, where the overlapped targets are extracted separately conditioned on the specific former prediction.
Nevertheless, most of the multi-stage methods suffer from error propagation.

\subsection{Tagging-based Information Extraction}
Tagging scheme in the field of information extraction has been extensively investigated.
Traditional sequence labeling approaches tagging each token once (e.g., \textit{BIO}) is hard to tackle irregular information extraction (e.g., overlapped NER).
Several researchers \cite{zheng2017joint} extend the BIO label scheme to adapt to more complex scenarios.
However, they suffer from the label ambiguity problem due to limited flexibility.
Recently, the grid tagging scheme is used in a lot of information extraction tasks, such as opinion mining \cite{wu2020grid}, relation extraction \cite{wang2020tplinker}, and named entity recognition \cite{wang2021discontinuous},
due to its characteristic of presenting relations between word pairs.
For example, TPLinker \cite{wang2020tplinker} realizes one-stage joint relation extraction without a gap between training and inference by tagging token pairs with link labels.
Inspired by these works, we design our tagging scheme to address overlapping and nested EE, which predicts relations between trigger or argument words parallelly in one stage.

Also it is noteworthy explicitly that this work inherits the recent success of the idea of word-word relation detection, as in \newcite{Li00WZTJL22}.
\newcite{Li00WZTJL22} propose to unify all the NER (including the flat, nested and discontinuous mentions) with a word-word modeling based on the grid tagging scheme.
This work however differs from \newcite{Li00WZTJL22} in two folds.
First, we extend the idea of the word-word tagging from NER to EE successfully, where we re-design two relation types for the nested and overlapped events.
Second, from the modeling perspective, we devise an adaptive event fusion layer to fully support the one-stage (end-to-end) complex event detection, which greatly helps avoid error propagation.

\section{Problem Formulation}

The goal of event extraction includes extracting event triggers and their arguments.
We can formalize overlapping and nested EE as follows:
given an input sentence consisting of N tokens or words $X = \{x_1, x_2, \dots, x_N\}$ and event type $e \in \mathcal{E}$, the task aims to extract the span relations $\mathcal{S}$ and the role relations $\mathcal{R}$ between each token pair $(x_i, x_j)$, where $\mathcal{E}$ denotes the event type collection, $\mathcal{S}$ and $\mathcal{R}$ are pre-defined tags.
These relations can be explained below, and we also give an example as demonstrated in Figure \ref{fig:label} for better understanding.
\begin{itemize}
\setlength{\topsep}{0pt}
\setlength{\itemsep}{0pt}
\setlength{\parsep}{0pt}
\setlength{\parskip}{0pt}
    \item $\mathcal{S}$: the span relation indicates that $x_i$ and $x_j$ are the starting and ending token of the extracted trigger span \texttt{S-T} or argument span \texttt{S-A}, where $1 \le i \le j \le N$.
    
    \item $\mathcal{R}$: the role relation indicates that the argument with $x_j$ acts the certain role \texttt{R-*} of the event with the trigger containing $x_i$, where $1 \le i, j \le N$. \texttt{*} indicates the role type.
    
    \item \texttt{NONE}, indicating that the word pair does not have any relation defined in this paper.
\end{itemize}

\section{Framework}

The architecture of our model is illustrated in Figure \ref{fig:architecture}, which mainly consists of three components. 
First, the widely-used pre-trained language model, BERT \cite{devlin2019bert}, is used as the encoder to yield contextualized word representations from the input sentences.
Then, an adaptive event fusion layer consisting of an attention module and two gate modules is used to integrate the target event type embedding into contextual representations.
Afterward, a prediction layer is employed to jointly extract the span relations and the role relations between word pairs.

\subsection{Encoder Layer}

We leverage BERT as the encoder for our model since it has been demonstrated to be one of the SoTA models for representation learning in EE.
Given an input sentence $X = \{x_1, x_2, \dots, x_N\}$, we convert each token $x_i$ into word pieces and then feed them into a pre-trained BERT module. 
After the BERT calculation, each sentential word may involve vectorial representations of several pieces. Here we employ max pooling to produce word representations $\bm{H} = \{\bm{h}_1, \bm{h}_2, ..., \bm{h}_N\} \in \mathbb{R}^{N \times d_h}$ based on the word piece representations.

\subsection{Adaptive Event Fusion Layer}

Since the goal of our framework is to predict the relations between word pairs for the target event type $e_t$, it is important to generate event-aware representations.
Therefore, to fuse the event information and contextual information provided by the encoder, we design an adaptive fusion layer.
As shown in Figure \ref{fig:architecture}, it consists of an attention module, modeling the interaction among events and obtaining the global event information, and two gate fusion modules for integrating the global and target event information with contextualized word representations.

\paragraph{Attention Mechanism}
Motivated by the self-attention in Transformer \cite{vaswani2017attention,NMCL}, we first introduce an attention mechanism, of which input consists of queries, keys, and values.
The output is computed as a weighted sum of the values, where the weight assigned to each value is the dot product of the query with the corresponding key.
The attention mechanism can be formulated as:
\begin{equation}
\setlength\abovedisplayskip{4pt}
\setlength\belowdisplayskip{4pt}
    {\rm Attention}(\bm{Q}, \bm{K}, \bm{V}) = {\rm softmax}(\frac{\bm{Q}\bm{K}^\top}{\sqrt{d_h}})\bm{V} \ ,
\end{equation}
where $\sqrt{d_h}$ is a scaling factor, $\bm{Q}$, $\bm{K}$ and $\bm{V}$ are query, key and value tensors, represented by Eq. \ref{eq:attn}.

\paragraph{Gate Fusion Mechanism}
We design a gate fusion mechanism to integrate two kinds of features and filter the unnecessary information.
The gate vector $\bm{g}$ is produced by a fully-connection layer with the sigmoid function, which can adaptively control the flow of the input:
\setlength\abovedisplayskip{4pt}
\setlength\belowdisplayskip{4pt}
\begin{align}
    {\rm Gate}(\bm{p}, \bm{q}) &= \bm{g} \odot \bm{p} + (1 - \bm{g}) \odot \bm{q} \ , \\
     \bm{g} &= \sigma(\bm{W}_g [\bm{p}; \bm{q}] + \bm{b}_g) \ ,
\end{align}
where $\bm{p}$ and $\bm{q}$ are input vectors, represented by Eq. \ref{eq:gate1} and Eq. \ref{eq:gate}. 
$\sigma(\cdot)$ is a sigmoid activation function, $\odot$ and $[;]$ denote element-wise product and concatenation operations, respectively. $\bm{W}_g$ and $\bm{b}_g$ are trainable parameters.

We leverage the attention mechanism to obtain the global event embeddings for each contextualized word representation.
Given a set of randomly initialized event type embeddings $\bm{E} = \{\bm{e}_1, \bm{e}_2, \dots, \bm{e}_M\} \in \mathbb{R}^{M \times d_h}$, where $M$ is the number of event types, the calculation can be formulated as:
\begin{equation}
    \bm{E}^g = {\rm Attention}(\bm{W}_q\bm{H}, \bm{W}_k\bm{E}, \bm{W}_v\bm{E}) \ ,
    \label{eq:attn}
\end{equation}
where $\bm{E}^g$ is the output of the attention mechanism, $\bm{W}_q$, $\bm{W}_k$ and $\bm{W}_v$ are learnable parameters.

To encode global event information into word representations, we adopt a gate module to fuse the contextual word representations and global event representations. 
After that, we employ another gate mechanism to integrate the target event type embedding and the output of the last gate module.
the overall process can be formulated as:
\setlength\abovedisplayskip{4pt}
\setlength\belowdisplayskip{4pt}
\begin{align}
\label{eq:gate1}
    \bm{H}^g &= {\rm Gate}(\bm{H}, \bm{E}^g) \ , \\
    \bm{V}^t &= {\rm Gate}(\bm{H}^g, \bm{e}_t) \ ,\ \label{eq:gate}
\end{align}
where $\bm{e}_t \in \bm{E}$ denotes the target event type embedding, $\bm{V}^t = \{\bm{v}_1, \bm{v}_2, ..., \bm{v}_N\} \in \mathbb{R}^{N \times d_h}$ is the final event-aware word representations.

\subsection{Joint Prediction Layer}

After the adaptive event fusion layer, we obtain the event-aware word representations $\bm{V}^t$, which are used to jointly predict the span and role relations between each pair of words.
For each word pair $(w_i, w_j)$, we calculate a score to measure the possibility of them for the relation $s \in \mathcal{S}$ and $r \in \mathcal{R}$.

\paragraph{Distance-aware Score} To integrate relative distance information and word pair representations, we introduce a distance-aware score function.
For two vectors $\bm{p}_i$ and $\bm{p}_j$ from a sequence of representations, we combine them with corresponding position embeddings from \newcite{su2021roformer}, and then calculate the score by the dot product of them:
\begin{equation}
\begin{aligned}
    {\rm Score}(\bm{p}_i, \bm{p}_j) &= (\bm{R}_i\bm{p}_i)^\top(\bm{R}_j\bm{p}_j) \\
    &=\bm{p}_i^\top\bm{R}_{j-i}\bm{p}_j \ ,
\end{aligned}
\end{equation}
where $\bm{R}_i$ and $\bm{R}_j$ are position embeddings of $\bm{p}_i$ and $\bm{p}_j$, $\bm{R}_{j-i} = \bm{R}_i^\top\bm{R}_j$.
Thus, we can obtain the span score $c^{s}_{ij}$ and the role score $c^{r}_{ij}$ of the word pair $(w_i, w_j)$ for target event type $t$:
\setlength\abovedisplayskip{4pt}
\setlength\belowdisplayskip{4pt}
\begin{align}
    c^{s}_{ij} &= {\rm Score}(\bm{W}_{s1}\bm{v}^t_i, \bm{W}_{s2}\bm{v}^t_j) \ , \\
    c^{r}_{ij} &= {\rm Score}(\bm{W}_{r1}\bm{v}^t_i, \bm{W}_{r2}\bm{v}^t_j) \ ,\
\end{align}
where $\bm{W}_{s1}$, $\bm{W}_{s2}$, $\bm{W}_{r1}$ and $\bm{W}_{r2}$ denote  parameters. $\bm{v}^t_i$ and $\bm{v}^t_j$ are from Eq. \ref{eq:gate}.

\subsection{Training Details}

For the score $c^{\star}_{ij}$, where $\star$ denotes the relation $s$ or $r$, our training target is to minimize a variant of circle loss \cite{sun2020circle} which extends softmax cross-entropy loss to figure out multi-label classification problem.
In addition, we introduce a threshold score $\delta$, noting that the scores of the pairs with relation are larger than $\delta$, and the other pairs are less than it. The loss function can be formulated as:
\begin{equation}
\mathcal{L}^\star = \log(e^{\delta} + \sum\limits_{(i,j)\in \Omega^\star} e^{-c^\star_{ij}}) + \log(e^{\delta} + \sum\limits_{(i,j)\notin \Omega^\star} e^{c^\star_{ij}}) \ , \\
\end{equation}
where $\Omega^\star$ denotes the pair set of relation $\star$, $\delta$ is set to zero. 

Finally, we enumerate all event types in the selected event type set $\mathcal{E}'$ and get the total loss:
\begin{equation}
    \mathcal{L} = \sum\limits_{t \in \mathcal{E}'} (\sum\limits_{s \in \mathcal{S}}\mathcal{L}^s + \sum\limits_{r \in \mathcal{R}}\mathcal{L}^r) \ , \\
\end{equation}
where $\mathcal{S}'$ is a subset sampled from $\mathcal{S}$, we detail the sampling strategy in the appendix.

\subsection{Inference}
\label{sec:inference}
\begin{figure}[t]
    \centering
    \includegraphics[width=1\columnwidth]{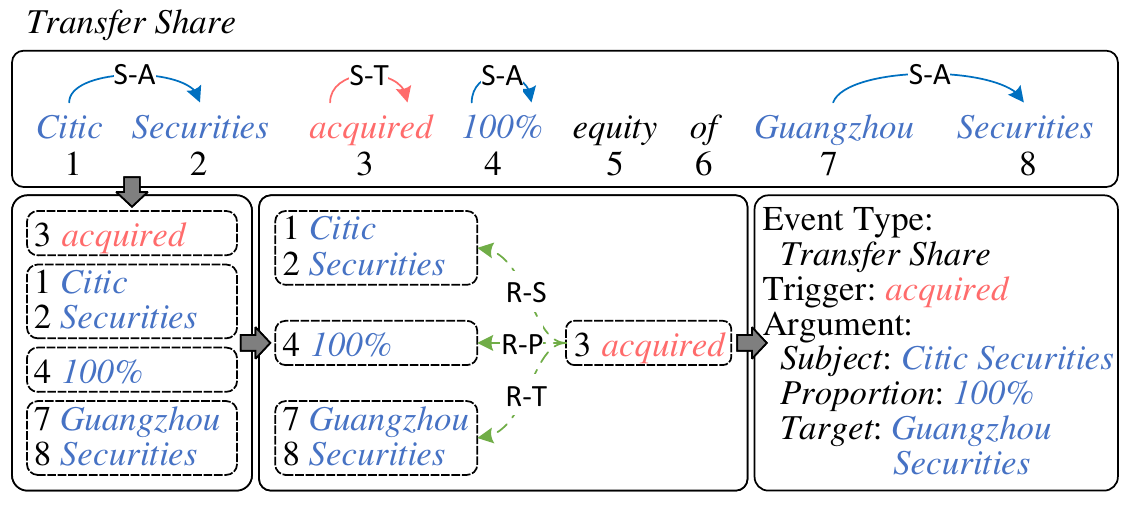}
    \caption{
    A decoding case of our system with four steps.
    }
    \label{fig:decode}
\end{figure}

During the inference period, our model is able to extract all events by parallelly injecting their event type embeddings to the adaptive event fusion layer. 
As shown in Figure \ref{fig:decode}, once all the tags of a certain event type are predicted by our model in one stage, 
the overall decoding process can be summarized as four steps:
First, we get starting and ending indices of the trigger or argument.
Second, we obtain the trigger and argument spans.\footnote{Note that if two pairs with the same span relation clash in the boundaries, the pair with higher score will be selected.}
Third, we match the trigger and arguments according to the R-\texttt{*} relations.
Finally, the event type is assigned to this event structure.
Specially, we repeat the above four steps for each event type.

\begin{table}[t]
\centering
\small
\begin{tabular}{llcccc}
\toprule
                            &       & \#Ovlp.   & \#Nest.    & \#Sent.   & \#Events \\ \midrule
\multirow{3}{*}{FewFC}      & train & 1,560      & -        & 7,185        & 10,227    \\
                            & dev   & 205       & -        & 899         & 1,281     \\
                            & test  & 210       & -        & 898         & 1,332     \\ \midrule
\multirow{3}{*}{Genia11} & train & 954       & 1,628     & 8,730        & 6,401     \\
                            & dev   & 121       & 199      & 1,091        & 824      \\
                            & test  & 125       & 197      & 1,092        & 775      \\ \midrule
\multirow{3}{*}{Genia13} & train & 347       & 784      & 4,000        & 2,743     \\
                            & dev   & 44        & 100      & 500         & 352      \\
                            & test  & 42        & 88       & 500         & 320      \\ \bottomrule
\end{tabular}
\caption{Statistics of the datasets. ``Ovlp.'' and ``Nest.'' denote the sentences with overlapped and nested events, respectively.}
\label{tab:dataset}
\end{table}

\begin{table*}[t!]
\centering
\small
\begin{tabular}{clcccccccccccc} 
 \toprule
 \multirow{2}{*}{} & \multirow{2}{*}{} & \multicolumn{3}{c}{TI(\%)} & \multicolumn{3}{c}{TC(\%)} & \multicolumn{3}{c}{AI(\%)}  & \multicolumn{3}{c}{AC(\%)} \\ \cmidrule(r){3-5} \cmidrule(r){6-8} \cmidrule(r){9-11} \cmidrule(r){12-14} 
  & & P & R & F1 & P & R & F1 & P & R & F1 & P & R & F1 \\  
  \midrule
  \multirow{3}{*}{\textbf{Flat EE}} &
BERT-softmax & 89.8 & 79.0 & 84.0 & 80.2 & 61.8 & 69.8 & 74.6 & 62.8 & 68.2 & 72.5 & 60.2 & 65.8 \\
& BERT-CRF & \textbf{90.8} & 80.8 & 85.5 & \textbf{81.7} & 63.6 & 71.5 & 75.1 & 64.3 & 69.3 & 72.9 & 61.8 & 66.9 \\
& BERT-CRF-joint & 89.5 & 79.8 & 84.4 & 80.7 & 63.0 & 70.8 & \textbf{76.1} & 63.5 & 69.2 &\textbf{74.2} & 61.2 & 67.1 \\
\cdashlinelr{1-14}
\multicolumn{1}{c}{\multirow{3}{*}{\begin{tabular}[c]{@{}c@{}}\textbf{Ovlp. \&}\\ \textbf{Nest. EE}\end{tabular}}} 
 &
PLMEE & 83.7 & 85.8 & 84.7 & 75.6 & 74.5 & 75.1 & 74.3 & 67.3 & 70.6 & 72.5 & 65.5 & 68.8 \\
\multicolumn{1}{c}{}   & MQAEE & 89.1 & 85.5 & 87.4 & 79.7 & 76.1 & 77.8 & 70.3 & 68.3 & 69.3 & 68.2 & 66.5 & 67.3 \\
\multicolumn{1}{c}{}   & CasEE & 89.4 & 87.7 & 88.6 & 77.9 & 78.5 & 78.2 & 72.8 & 73.1 & 72.9 & 71.3 & 71.5 & 71.4 \\ \cdashlinelr{1-14}
{\textbf{Ours}} &
OneEE & 88.7 & \textbf{88.7} & \textbf{88.7} & 79.1 & \textbf{80.3} & \textbf{79.7} & 75.4 & \textbf{77.0} & \textbf{76.2} & 74.0 & \textbf{72.9} & \textbf{73.4} \\  
\bottomrule
\end{tabular}
\caption{Results for extracting all kinds of events on FewFC, where TI, TC, AI, AC denote trigger identification, trigger classification, argument identification, and argument classification, respectively. We run our model for 5 times with different random seeds and report the median values.}
\label{tab:fewfc_table}
\end{table*}

\section{Experiments Settings}

\subsection{Datasets}
As shown in Table \ref{tab:dataset}, we follow previous work \cite{sheng2021casee}, adopting FewFC \cite{zhou2021role}, a Chinese financial event extraction benchmark for overlapped EE.
FewFC annotates 10 event types and 18 argument role classes with about 22\% sentences containing overlapped events.
We also experiment on two biomedical datasets for nested EE, namely Genia11 \cite{kim2011overview} and Genia13 \cite{kim2013Genia},
with around 18\% sentences containing nested events.
Genia11 annotates 9 event types and 10 argument role classes while the figures for Genia13 are 13 and 7.
We split the train/dev/test as 8.0:1.0:1.0 for both of them.

\subsection{Implementation Details}
We employ the Chinese Bert-base model for FewFC and BioBERT \cite{lee2020biobert} for Genia11 and Genia13. 
We adopt AdamW \cite{loshchilov2019decoupled} optimizer with the learning rate of $2e-5$ for BERT and $1e-3$ for the other modules. 
The batch size is 8 and the hidden size $d_h$ is 768.
We train our model with 20 epochs on FewFC and Genia11 and 30 epochs on Genia13.
All the hyper-parameters are tuned on the development set.
All the event type embeddings are trained from scratch.

\subsection{Evaluation Metrics}
For evaluation, we follow the traditional criteria of previous work \cite{chen2015event,du2020event,sheng2021casee}.
1) Trigger Identification (TI): A trigger is correctly identified if the predicted trigger span matches with a golden label;
2) Trigger Classification (TC): A trigger is correctly classified if it is correctly identified and assigned to the right type;
3) Argument Identification (AI): An argument is correctly identified if its event type is correctly recognized and the predicted argument span matches with a golden label;
4) Argument Classification (AC): An argument is correctly classified if it is correctly identified and the predicted role matches any of the golden labels.
We report Precision (P), Recall (R), and F measure (F1) for each of the four metrics.

\subsection{Baselines}

\paragraph{Sequence Labeling Methods for Flat EE}
These methods cast the EE task into a sequence labeling task by assigning each token a label.
\textbf{BERT-softmax} uses BERT to get feature representations for classifying triggers and arguments.
\textbf{BERT-CRF} adds the CRF layer on BERT to capture label dependencies.
\textbf{BERT-CRF-joint} extends the BIO tagging scheme to joint labels of type and role as \texttt{B/I/O-type-role}, inspired by joint extraction of entity and relation \cite{zheng2017joint}.
All these methods are incapable to solve the overlapping problem due to label conflicts.
\paragraph{Multi-stage Methods for Overlapped and Nested EE}
These methods perform EE in several stages. \textbf{PLMEE} \cite{yang2019exploring} solves the argument overlap problem by extracting role-specific argument according to the trigger predicted by the trigger extractor in a pipeline manner.
\textbf{CasEE} \cite{sheng2021casee} sequentially performs type\&trigger\&argument extractions, where the overlapped targets are separately extracted conditioned on former predictions and all subtasks are jointly learned.

\section{Experimental Results}

\subsection{Results of All EE}
Table \ref{tab:fewfc_table} reports the result of all methods on the overlapped EE dataset, FewFC, while Table \ref{tab:Genia_table} reports the results of the nested EE datasets, Genia11 and Genia13.
We can observe that:

1) Our method significantly outperforms all other methods and achieves the state-of-the-art F1 score on all three datasets.

2) In comparison with sequence labeling methods, our model achieves better recall and F1 scores.
Specifically, our model outperforms BERT-CRF-joint by 11.7\% and 6.3\% in recall and the F1 score of AC on the FewFC dataset and achieves a substantial improvement of 4.4\% in F1 score of AC on two Genia datasets averagely.
It shows the effectiveness of our model on overlapped and nested EE since the sequence labeling methods can only solve flat EE.

3) In comparison with multi-stage methods, our model also improves the performance on the F1 score considerably.
Our model outperforms the state-of-the-art model, CasEE, by 2.1\% in the F1 score of TC on three datasets averagely.
We consider this is because that the event feature is well learned by our adaptive event fusion module.
Especially, our model improves 3.4\% on AI and 1.6\% on AC over CasEE on an average of three datasets.
The results reveal the superiority of our one-stage framework which elegantly realizes overlapped and nested event extraction without error propagation.

\begin{table}[t]
\centering
\small
\begin{tabular}{lcccc}
\toprule
      & TI(\%) & TC(\%) & AI(\%) & AC(\%) \\ \midrule
\multicolumn{5}{l}{$\bullet$ \bf Genia11}\\ \cdashlinelr{1-5}
BERT-softmax & 67.8 & 64.4 & 57.4 & 56.0 \\
BERT-CRF & 68.3 & 64.8 & 58.3 & 56.9 \\
BERT-CRF-joint & 67.0 & 64.1 & 60.2 & 58.1 \\ \cdashlinelr{1-5}
PLMEE & 67.3 & 65.5 & 60.7 & 59.4 \\
CasEE & 70.0 & 67.0 & 62.0 & 60.4 \\ \cdashlinelr{1-5}
OneEE  & \textbf{71.5} & \textbf{69.5} & \textbf{65.9} & \textbf{62.5} \\ \midrule \midrule
\multicolumn{5}{l}{$\bullet$ \bf Genia13}\\ \cdashlinelr{1-5}
BERT-softmax & 77.4 & 75.9 & 69.9 & 67.7 \\
BERT-CRF & 78.8 & 77.4 & 70.1 & 68.2 \\
BERT-CRF-joint & 77.6 & 75.7 & 71.9 & 68.2 \\ \cdashlinelr{1-5}
PLMEE & 79.3 & 78.3 & 72.1 & 70.7 \\
CasEE & 80.5 & 78.5 & 73.7 & 71.9 \\ \cdashlinelr{1-5}
OneEE  & \textbf{81.9} & \textbf{80.8} & \textbf{76.8} & \textbf{72.7} \\ \bottomrule

\end{tabular}
\caption{F1 scores for extracting all events on Genia11 and Genia13. 
}
\label{tab:Genia_table}
\end{table}

\begin{figure}[t]
\centering
\small
\begin{tikzpicture}
  \begin{axis}[name=plot1,ybar=7pt,height=0.45\columnwidth,width=0.57\columnwidth,ylabel=F1 (\%) on FewFC,xlabel=(a) Overlapped TC,xtick style={draw=none},enlargelimits=0.5,xticklabel style = {yshift=5pt},ylabel shift=-3pt,xticklabels={,,},compat=newest,
  legend style={at={(1.03,1.33)},draw=none, font=\small,
    anchor=north,legend columns=4,
        /tikz/every even column/.append style   = {column sep=0.1cm},
                 /tikz/every odd column/.append style    = {column sep=0.1cm,}
                 }
                 ,xlabel style={at={(0.45,-0.05)}}]
    \addplot[blue,fill=blue!30!white] coordinates {(0,52.0)};
    \addplot[black,fill=gray] coordinates {(0,66.6)};
    \addplot[brown!60!black,fill=brown!30!white] coordinates {(0,74.9)};
    \addplot[red,fill=red!30!white] coordinates {(0,77.79)};
    \legend{BERT-CRF-joint,PLMEE, CasEE,Ours}
  \end{axis}
  \begin{axis}[name=plot2,at={($(plot1.east)+(0.7cm,0)$)},anchor=west,ybar=7pt,height=0.45\columnwidth,width=0.57\columnwidth,xtick style={draw=none},enlargelimits=0.5,xticklabel style = {yshift=5pt},ylabel shift=-5pt,xticklabels={,,},compat=newest,xlabel=(b) Overlapped AC,xlabel style={at={(0.45,-0.05)}}]
    \addplot[blue,fill=blue!30!white] coordinates {(0,56.8)};
    \addplot[black,fill=gray] coordinates {(0,61.4)};
    \addplot[brown!60!black,fill=brown!30!white] coordinates {(0,70.3)};
    \addplot[red,fill=red!30!white] coordinates {(0,72.8)};
  \end{axis}
  \begin{axis}[name=plot3,at={($(plot1.south)-(0,0.7cm)$)},anchor=north,ybar=7pt,height=0.45\columnwidth,ylabel=F1 (\%) on Genia11,width=0.57\columnwidth,xtick style={draw=none},enlargelimits=0.5,xticklabel style = {yshift=5pt},ylabel shift=-5pt,xticklabels={,,},compat=newest,xlabel=(c) Nested TC,xlabel style={at={(0.45,-0.05)}}]
    \addplot[blue,fill=blue!30!white] coordinates {(0,69)};
    \addplot[black,fill=gray] coordinates {(0,72.8)};
    \addplot[brown!60!black,fill=brown!30!white] coordinates {(0,74.5)};
    \addplot[red,fill=red!30!white] coordinates {(0,76.11)};
  \end{axis}
  \begin{axis}[name=plot4,at={($(plot2.south)-(0,0.7cm)$)},anchor=north,ybar=7pt,height=0.45\columnwidth,width=0.57\columnwidth,xtick style={draw=none},enlargelimits=0.5,xticklabel style = {yshift=5pt},ylabel shift=-5pt,xticklabels={,,},compat=newest,xlabel=(d) Nested AC,xlabel style={at={(0.45,-0.05)}}]
    \addplot[blue,fill=blue!30!white] coordinates {(0,65)};
    \addplot[black,fill=gray] coordinates {(0,66.8)};
    \addplot[brown!60!black,fill=brown!30!white] coordinates {(0,69.0)};
    \addplot[red,fill=red!30!white] coordinates {(0,69.47)};
  \end{axis}
\end{tikzpicture}
\caption{Results for overlapped trigger (a) and argument (b) extraction on FewFC, and nested trigger (c), and argument (d) extraction on Genia11. Note that only the sentences that contain at least one such event are used.
}
\label{fig:overlap_analysis}
\end{figure}
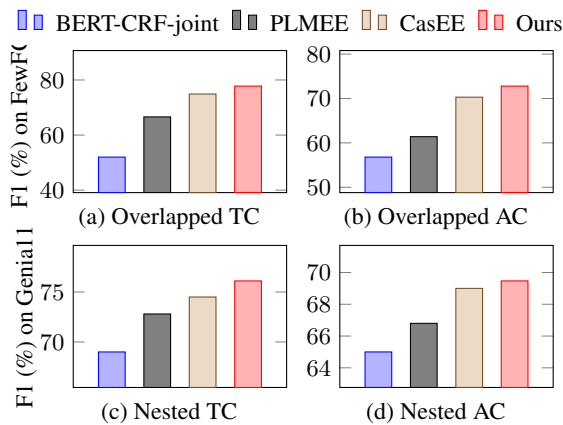

\subsection{Results of Overlapped and Nested EE}
To evaluate the effectiveness of our proposed model in recognizing overlapping and nested event mentions, we further report the results on sentences containing at least one overlapping event in FewFC and sentences containing at least one nested event in Genia11, respectively.

Figure \ref{fig:overlap_analysis} illustrates the results of TC and AC on overlapping and nested sentences in testing.
It shows that our method outperforms other methods on overlapping and nested sentences.
The reasons are mainly two-fold:
1) We solve all the overlapping patterns while BERT-CRF-joint could not handle overlapped and nested EE and PLMEE only solve the argument overlap.
2) Our one-stage model outperforms CasEE because we effectively learn event-aware representations and extract word-word relations in parallel, while CasEE performs in three sequential steps with error propagation.

\subsection{Effects of the Modules in the Fusion Layer}

\begin{table}[t]
\centering
\small
\begin{tabular}{lcccc}
\toprule
      & TI(\%) & TC(\%) & AI(\%) & AC(\%) \\ \midrule
OneEE   & 88.7 & 79.7 & 76.2 & 73.4 \\ \cdashlinelr{1-5}
w/o Attention & 88.3 & 79.5 & 75.9 & 72.8  \\
w/o Gate & 88.4 & 79.3 & 75.3 & 72.6  \\
w/o Fusion Layer & 88.0 & 78.7 & 75.2 & 72.2 \\ \cdashlinelr{1-5}
w/o Position Emb. & 88.1 & 78.7 & 74.1 & 71.8 \\ \bottomrule

\end{tabular}
\caption{Ablation studies using FewFC.}
\label{tab:ablation}
\end{table}

To verify the effectiveness of each component, we conduct ablation studies on the FewFC dataset, as shown in Table \ref{tab:ablation}.
First, without the attention mechanism, we observe slight performance drops.
By replacing the gate mechanism with an addition operation, the performance also decreases to a small degree.
Furthermore, a significant drop can be found when the adaptive event fusion layer is substituted by addition, which indicates the usefulness of event representation and context.
Finally, removing the position embeddings results in a remarkable drop on all F1 scores, especially 1.6\% in the F1 score of AC, which suggests that the information of positions is essential to recognize word-word relations.

\subsection{Effect of the Distance-aware Tag Prediction}

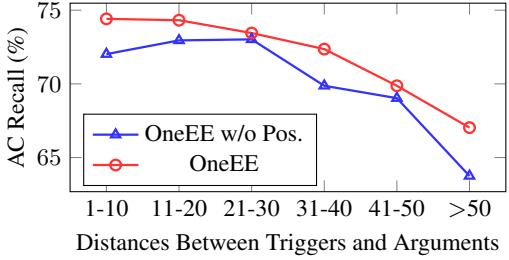
\begin{figure}[t]
\hspace{-2mm}
    \centering
    \small
    \begin{tikzpicture}
	\begin{axis}[
		xlabel=Distances Between Triggers and Arguments,
		ylabel=AC Recall (\%),
		ylabel shift=-3pt,
		compat=newest,
		xtick=data,
		symbolic x coords={1-10, 11-20, 21-30, 31-40, 41-50, $>$50},
		legend pos=south west,
		height=0.53\columnwidth,width=0.95\columnwidth,
]

	\addplot[color=blue!80!white, mark=triangle, line width=0.3mm] coordinates {
		(1-10, 72.02)
		(11-20, 72.95)
		(21-30, 73.02)
		(31-40, 69.87)
		(41-50, 69.03)
        ($>$50, 63.75)
	};
	\addplot[color=red!80!white, mark=o, line width=0.3mm] coordinates {
		(1-10, 74.41)
		(11-20, 74.32)
		(21-30, 73.45)
		(31-40, 72.36)
		(41-50, 69.87)
        ($>$50, 67.03)
	};
	\legend{OneEE w/o Pos. , OneEE}
	\end{axis}
\end{tikzpicture}
    \caption{FewFC results of extracting triggers and arguments with different distances. The red line denotes that position embeddings are used while the blue line not.
    }
    \label{fig:distance}
\end{figure}

In this section, we investigate the effect of position embeddings for the prediction layer of OneEE. 
We divide the arguments in the test set of FewFC into 6 groups according to their distance from corresponding triggers and report the recall scores of the model with and without position embeddings.
As shown in Figure \ref{fig:distance}, 
the AC recall declines as the distance between trigger and argument in an event go up.
This indicates that it is more difficult for the model to detect roles correctly if the distance is longer in an event.
Furthermore, the model with position embeddings outperforms another one without position embeddings,
revealing that the relative distance information is beneficial for event extraction.

\begin{table}[t]
\centering
\small
\begin{tabular}{lcccc}
\toprule
Model & Stage & \#Param. & Speed (sent/s) & Ratio \\ \midrule
PLMEE                  & Two & 204.6M & 19.8   & $\times$1.0      \\              
CasEE                  & Three & 120.7M & 62.3 & $\times$3.1   \\         
OneEE                   & One &  114.2M & 79.4 &  $\times$4.0   \\ 
OneEE$^\dag$         & One &  114.2M & 186.5  & $\times$9.4   \\ \bottomrule
\end{tabular}
\caption{Parameter number and inference speed comparisons on FewFC. All models are tested with batch size 1, $\dag$ denotes that the model is tested with batch size 8. The ratio denotes the multiple of the speed increase with regard to PLMEE.}
\label{tab:speed}
\end{table}

\subsection{Parameter Number \& Efficiency Comparisons}
Table \ref{tab:speed} lists the stage numbers, parameter numbers, and inference speeds of two baselines and our model.
For a fair comparison, all of these models are implemented using PyTorch and tested using the NVIDIA RTX 3090 GPU, where the batch size is set as 1.
As seen, PLMEE has 2 times as many parameters as the other two models, due to the utilization of two BERT-based modules for each stage.
Moreover, the inference speed of our model is about 3 times faster than that of PLMEE \cite{yang2019exploring} and 0.3 times faster than that of CasEE \cite{sheng2021casee}, which verifies the efficiency of our model.
Last but not least, when the batch size is set as 8, the inference speed of our model is 9.4 times as fast as that of PLMEE,
which also demonstrates the advantage of our model, that is, it supports parallel inference.
In one word, our model leverages fewer parameters but achieves better performance and faster inference speed.

\begin{figure}[t]
     \centering
     \subfigure[TH-AH]{
         \centering
         \includegraphics[width=0.45\columnwidth]{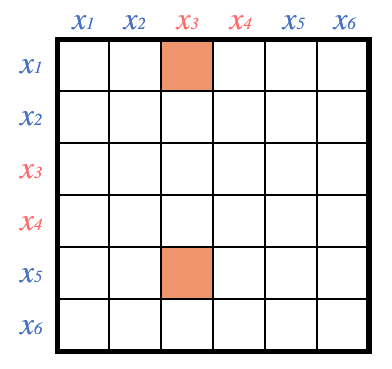}
         \label{fig:role_label1}
     }
     \subfigure[TW-AH]{
         \centering
         \includegraphics[width=0.45\columnwidth]{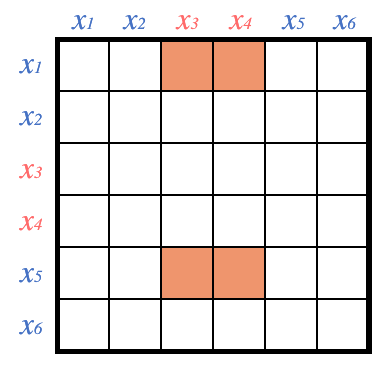}
         \label{fig:role_label2}
     }
     \subfigure[TH-AW]{
         \centering
         \includegraphics[width=0.45\columnwidth]{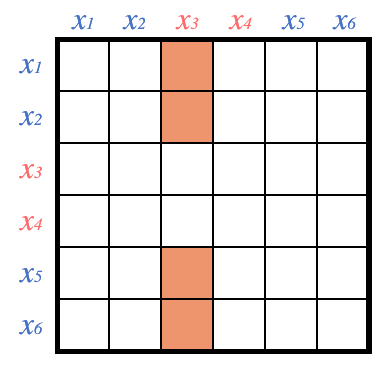}
         \label{fig:role_label3}
     }
     \subfigure[TW-AW]{
         \centering
         \includegraphics[width=0.45\columnwidth]{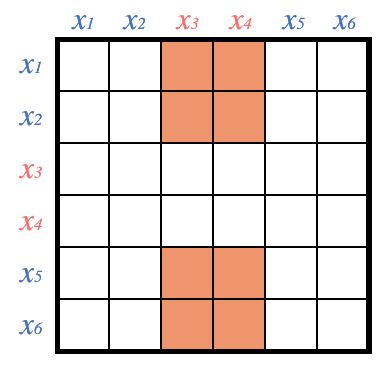}
         \label{fig:role_label4}
     }
    \caption{Four kinds of role label strategies. The goal is to predict the relation between trigger head and argument head (a), trigger word and argument head (b), trigger head and argument word (c), and trigger word and argument word (d).}
    \label{fig:role_label}
\end{figure}

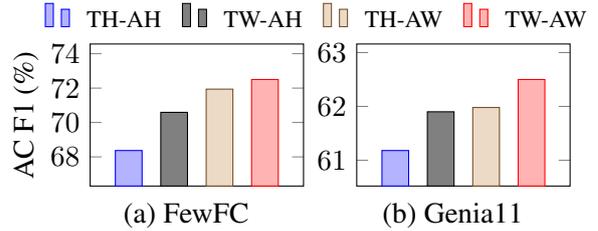
\begin{figure}[t]
\centering
\begin{tikzpicture}
  \begin{axis}[name=plot1,ybar=7pt,height=0.45\columnwidth,width=0.57\columnwidth,ylabel=AC F1 (\%),xlabel=(a) FewFC,xtick style={draw=none},enlargelimits=0.5,xticklabel style = {yshift=5pt},ylabel shift=-3pt,xticklabels={,,},compat=newest,
  legend style={at={(1.07,1.33)},draw=none, font=\small,
        anchor=north,legend columns=4,
        /tikz/every even column/.append style   = {column sep=0.2cm},
                 /tikz/every odd column/.append style    = {column sep=0.15cm,}
                 }
                 ,xlabel style={at={(0.45,-0.05)}}]
    \addplot[blue,fill=blue!30!white] coordinates {(0,68.37)};
    \addplot[black,fill=gray] coordinates {(0,70.59)};
    \addplot[brown!60!black,fill=brown!30!white] coordinates {(0,71.94)};
    \addplot[red,fill=red!30!white] coordinates {(0,72.5)};
    \legend{TH-AH, TW-AH, TH-AW, TW-AW}
  \end{axis}
  \begin{axis}[name=plot2,at={($(plot1.east)+(0.7cm,0)$)},anchor=west,ybar=7pt,height=0.45\columnwidth,width=0.57\columnwidth,xtick style={draw=none},enlargelimits=0.5,xticklabel style = {yshift=5pt},ylabel shift=-5pt,xticklabels={,,},compat=newest,xlabel=(b) Genia11,xlabel style={at={(0.45,-0.05)}}]
    \addplot[blue,fill=blue!30!white] coordinates {(0,61.18)};
    \addplot[black,fill=gray] coordinates {(0,61.90)};
    \addplot[brown!60!black,fill=brown!30!white] coordinates {(0,61.98)};
    \addplot[red,fill=red!30!white] coordinates {(0,62.5)};
  \end{axis}
\end{tikzpicture}
\caption{Results of AC with different role label strategies on FewFC (a) and Genia11 (b) datasets.
}
\label{fig:role_strategy}
\end{figure}

\subsection{Analysis of 4 Role Label Strategies}

In this section, we investigate the effect of the role strategies for AC performance. As shown in Figure \ref{fig:role_label}, we introduce 4 different strategies to predict the role relation between trigger and argument: the role labels only exist in 1) trigger and argument head pairs (TH-AH), 2) trigger word and argument head pairs (TW-AH), 3) trigger head and argument word pairs (TH-AW), and 4) trigger and argument word pairs (TW-AW).
The results of our model with 4 strategies are demonstrated in Figure \ref{fig:role_strategy}. 
We can learn that TW-AW achieves the best results against all other strategies on both FewFC and Genia11 datasets.
It is largely due to that its labels are denser than other strategies.

\subsection{Analysis of the Event Number}

\begin{figure}[t]
\centering
\begin{tikzpicture}
\begin{axis} [
ybar, 
ylabel=TC F1 (\%), 
width=1.05\columnwidth, 
height=0.5\columnwidth, 
xtick style={draw=none}, 
enlargelimits=0.4,
bar width=9pt,
extra y ticks={0}, 
y tick label style={font=\small}, 
symbolic x coords={1, 2, $>$2},
xtick=data,
xticklabel style = {yshift=5pt},
ylabel shift=-3pt, 
  legend style={at={(0.45,1.28)},draw=none, font=\small,
        anchor=north,legend columns=4,
        /tikz/every even column/.append style   = {column sep=0.1cm},
                 /tikz/every odd column/.append style    = {column sep=0.1cm,}
                 },
compat=newest] 
\addplot[blue,fill=blue!30!white] coordinates {(1, 78.9) (2, 60.6) ($>$2, 54.5)
};
\addplot[black,fill=gray] coordinates {(1, 78.7) (2, 74.5) ($>$2, 67.6)
};
\addplot[brown!60!black,fill=brown!30!white] coordinates {(1, 78.2) (2, 80.6) ($>$2, 70.3)
};
\addplot[red,fill=red!30!white] coordinates {(1, 80.63) (2, 81.61) ($>$2, 71.32)
};
\legend{BERT-CRF-joint, PLMEE, CasEE, OneEE}
\end{axis}
\end{tikzpicture}
\begin{tikzpicture}
\begin{axis} [
ybar, 
xlabel=Event Number,
ylabel=AC F1 (\%), 
width=1.05\columnwidth, 
height=0.5\columnwidth, 
xtick style={draw=none}, 
enlargelimits=0.4,
bar width=9pt,
extra y ticks={0}, 
y tick label style={font=\small}, 
symbolic x coords={1, 2, $>$2},
xtick=data, 
xlabel shift=-3pt,
ylabel shift=-3pt, 
xticklabel style = {yshift=5pt},
compat=newest] 
\addplot[blue,fill=blue!30!white] coordinates {(1, 71.4) (2, 56.8) ($>$2, 47.2)
};
\addplot[black,fill=gray] coordinates {(1, 70.5) (2, 68.9) ($>$2, 58.1)
};
\addplot[brown!60!black,fill=brown!30!white] coordinates {(1, 70.3) (2, 73.8) ($>$2, 61.3)
};
\addplot[red,fill=red!30!white] coordinates {(1, 74.55) (2, 75.86) ($>$2, 62.09)
};
\end{axis}
\end{tikzpicture}
\caption{Results of different event numbers on FewFC.}
\label{fig:event_number}
\end{figure}
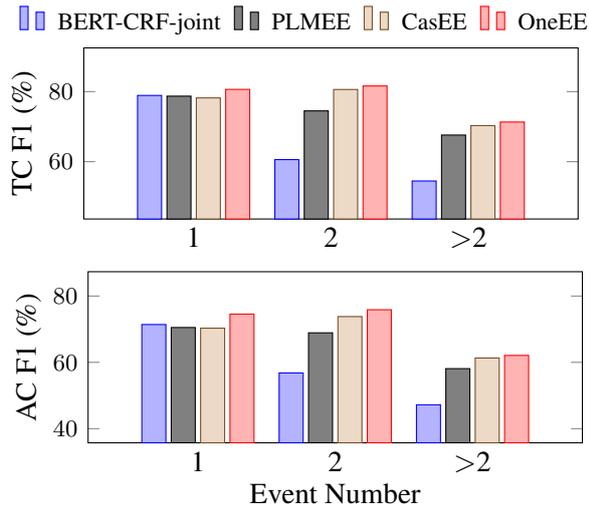

We further investigate the effect of the event number for EE, and the results are shown in Figure \ref{fig:event_number}.
We can observe that BERT-CRF-joint, PLMEE, and CasEE achieve similar performances on single-event sentences, while CasEE outperforms PLMEE and BERT-CRF-joint on the sentences with multiple events.
Most importantly, our system achieves the best results against all other baselines for different event numbers,
indicating the advances of our proposed method.

\section{Conclusion}
In this paper, we propose a novel one-stage framework based on word-word relation recognition to address overlapped and nested EE concurrently.
The relations between word pairs are pre-defined as the word-word relations within a trigger or argument and cross a trigger-argument pair.
Moreover, we propose an efficient model that consists of an adaptive event fusion layer for integrating the target event representation, and a distance-aware prediction layer for identifying all kinds of relations jointly. 
Experimental results show that our proposed model achieves new SoTA results on three datasets and faster speed than the SoTA model.
Through ablation studies, we find that the adaptive event fusion layer and distance-aware prediction layer are effective in improving the model performance.
In future work, we will extend our method to other structured prediction tasks, such as structured sentiment analysis and overlapped entity relation extraction.

\section*{Acknowledgements}

This work is supported by the National Natural Science Foundation of China (No. 62176187), the National Key Research and Development Program of China (No. 2017YFC1200500), the Research Foundation of Ministry of Education of China (No. 18JZD015), the Youth Fund for Humanities and Social Science Research of Ministry of Education of China (No. 22YJCZH064), the General Project of Natural Science Foundation of Hubei Province (No.2021CFB385). L Zhao would like to thank the support from Center for Artificial Intelligence (C4AI-USP), the Sao Paulo Research Foundation (FAPESP grant \#2019/07665-4), the IBM Corporation, and China Branch of BRICS Institute of Future Networks.


\clearpage
\newpage

\appendix

\section{Parallel Training with Sampling}

We parallelly inject multiple target event type embeddings at the adaptive event fusion layer during training period, which results in huge computation resources.
To this end, we use a subset $\mathcal{E}'$ to replace $\mathcal{E}$ for each sample, where the number of $\mathcal{E}'$ is $K$.
It consists of one positive event type (the event type annotated in the sample) and $K-1$ negative event types selected randomly from the event types that does not appear in the sample. 
In other words, we inject $K$ different event type embeddings into the gate module of Eq. \ref{eq:gate} simultaneously.
If there is no positive event type in the sample, we will select $K$ negative event types.

\section{Decoding for Nested EE}

\begin{figure}[h]
     \centering
     \subfigure[Nested Event Example]{
         \centering
         \includegraphics[width=0.95\columnwidth]{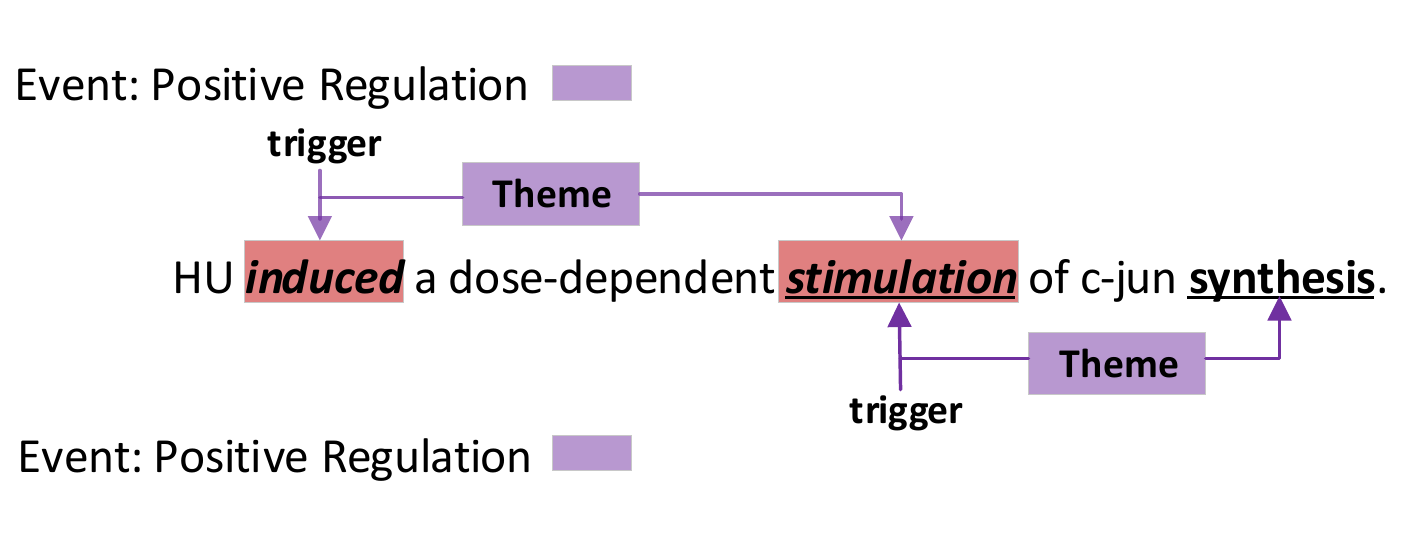}
     }
     \subfigure[Nested EE Decoding]{
         \centering
         \includegraphics[width=1\columnwidth]{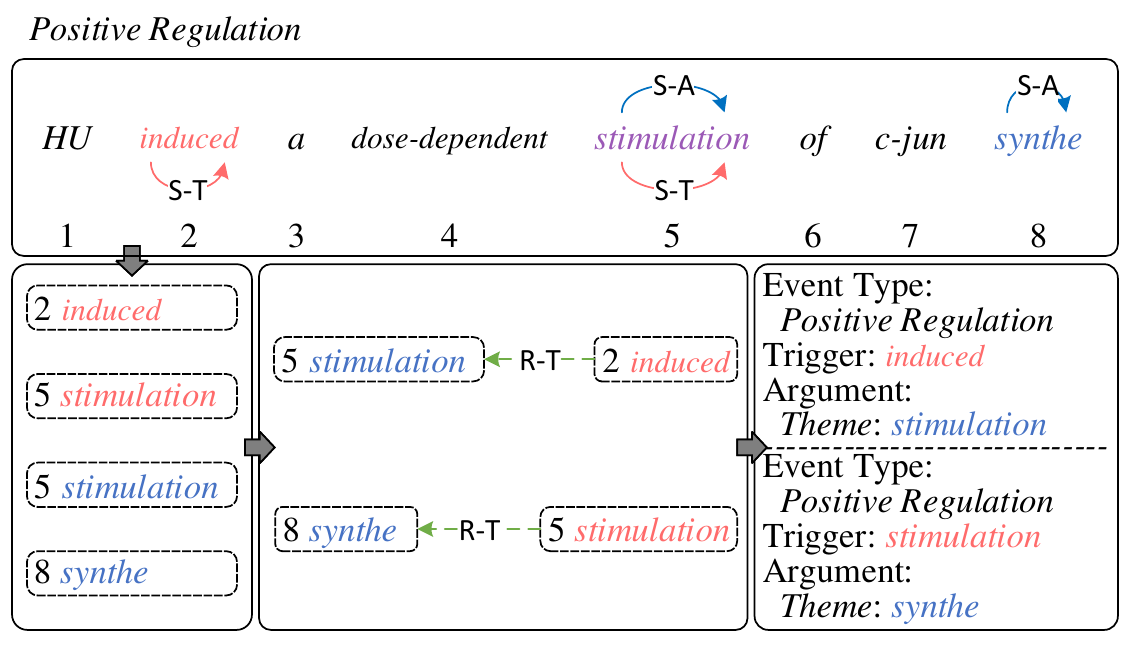}
     }
    \caption{Example of nested event (a) and its decoding process (b).
    }
    \label{fig:decode_nest}
\end{figure}

In the manuscript, we have already shown the decoding process of our model for overlapped EE in Section \ref{sec:inference}. 
Due to page limitation, we show an example of nested in Figure \ref{fig:decode_nest}(a).
We also demonstrate its decoding process in Figure \ref{fig:decode_nest}(b), which is the same as the overlapped EE decoding.

\section{Analysis of the Event Sampling Number}

\begin{figure}[t]
    \centering
    \begin{tikzpicture}
	\begin{axis}[
		xlabel=Sampling Number $K$,
		ylabel=TC F1 (\%),
		ylabel shift=-3pt,
		compat=newest,
		xtick=data,
		legend pos=south east,
		height=0.5\columnwidth,width=0.99\columnwidth]
	\addplot[color=blue!80!white, mark=triangle, line width=0.3mm] coordinates {
		(2, 76.51)
		(3, 77.79)
		(4, 78.86)
		(5, 79.28)
		(6, 79.43)
		(7, 79.31)
		(8, 78.93)
		(9, 78.81)
	};
	\addplot[color=red!80!white, mark=o, line width=0.3mm] coordinates {
		(2, 76.79)
		(3, 77.98)
		(4, 79.10)
		(5, 79.38)
		(6, 79.7)
		(7, 79.24)
		(8, 78.89)
		(9, 78.65)
	};
	\legend{Random, Pos. and Neg.}
	\end{axis}
\end{tikzpicture}
    \caption{Results on different sampling numbers of random sampling, and positive and negative sampling.}
    \label{fig:sampling}
\end{figure}
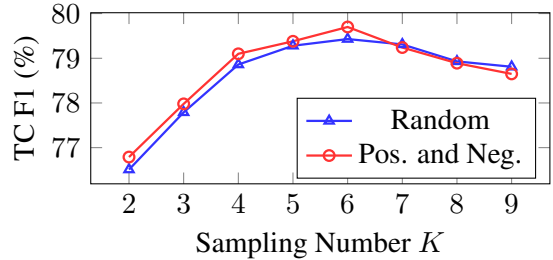

To further analyze the effect of sampling number $K$ and the sampling strategy, 
we also evaluate our model with positive and negative sampling and random sampling and compare them with different sampling numbers.
Figure \ref{fig:sampling} shows the TC F1 change trend as the number of sampling increases.
As seen, both two models with 6 event type samplings achieve the best performance, compared with the other sampling numbers.
Specifically, our model with one positive sampling and $K-1$ negative samplings outperforms the model with $K$ randomly selected samplings when $K$ is less than 7,
which demonstrates that our sampling strategy is helpful for the model training.

\end{CJK}
\end{document}